# PROSODY BASED CO-ANALYSIS FOR CONTINOUS RECOGNITION OF CO-VERBAL GESTURES

*Sanshzar Kettebekov, Mohammed Yeasin, Rajeev Sharma*

Department of Computer Science and Engineering
Pennsylvania State University
220 Pond Laboratory
University Park, PA 16802, USA
[kettebek; yeasin; rsharma]@cse.psu.edu

## ABSTRACT

This paper presents a Bayesian approach that uses both audio and visual cues for automatic recognition of continuous co-verbal gestures. The prosodic features from the speech signal was co-analyzed with the visual signal to learn the prior probability of occurrence of a particular class of gesture ($\Rightarrow$). The visual signal was used to compute the likelihood of $\Rightarrow$. It was found that the above co-analysis helps in detecting substitution and deletion which subsequently improves the rate of continuous gesture recognition. The efficacy of the proposed approach was demonstrated on a large database collected from the weather channel broadcast.

## 1. INTRODUCTION

In combination, gesture and speech constitute the most important modalities in human-to-human communication. People use large variety of gestures either to convey what cannot always be expressed using speech only or to add expressiveness to the communication. Motivated by this there has been a considerable interest in incorporating both gestures and speech as the means for Human-Computer Interaction (HCI).

To date, speech and gesture recognition have been studied extensively but most of the attempts at combining them in an interface were in the form of a predefined signs and controlled syntax such as "*put <point> that <point> there*", e.g., [1]. Part of the reason for the slow progress in multimodal HCI is the lack of available sensing technology that would allow non-invasive acquisition of natural behavior. However, the availability of abundant processing power has contributed to making computer vision based continuous gesture recognition in real time to allow the inclusion of natural gesticulation in a multimodal interface [2-4].

State of the art in *continuous gesture recognition* is far from requirements of a multimodal HCI due to poor recognition rates. Co-analysis of visual gesture and speech signals provide an attractive prospect of improving continuous gesture recognition. However, lack of fundamental understanding of speech/gesture production mechanism restricted implementation of the multimodal integration at the semantic level, e.g. [3-5]. Previously, we showed somewhat significant improvement in co-verbal gesture recognition when those were co-analyzed with keywords [3]. However, the implications of using top-down approach has augmented challenges with those of natural language and gesture processing and made automatic processing challenging.

The goal of the present work is to investigate co-occurrence of speech and gesture as applied to continuous gesture recognition from a bottom-up perspective. Instead of keywords, we employ a set of prosodic features from speech that correlate with co-verbal gestures. We address the general problem in multimodal HCI research, e.g., availability of valid data, by using narration sequences from the weather channel TV broadcast. The paper organized as follows. First we review types of gestures that are suitable for HCI and underline issues of addressing recognition of co-verbal gestures. Section 3 presents a framework for correlating visual and speech signals. Section 4 shows integration of prosodic features for gesture recognition and discusses results.

## 2. CO-VERBAL GESTURES FOR HCI

McNeill [6] distinguishes four major types of gestures by their relationship to the speech. *Deictic* gestures are used to direct a listener's attention to a physical reference in course of a conversation. These gestures, mostly limited to the pointing, were found to be co-verbal, cf. [6]. From our previous studies, in the computerized map domain (*i*MAP) [7], over 93% of deictic gestures were observed to co-occur with spoken nouns, pronouns, and spatial adverbials.

*Iconic* and *metaphoric* gestures are associated with abstract ideas, mostly peculiar to subjective notions of an individual. *Beats* serve as gestural marks of speech pace. In the weather channel broadcast the last three categories roughly constitute 20% of all the gestures exhibited by the narrators. We limit our current study to the *deictic* gestures as being more suitable for HCI.

## 2.1. Kinematics of Continuous Gestures

A continuous hand gesture consists of a series of qualitatively different kinematical phases such as movement to a position, hold, and transitional movement. We adopt Kendon's framework [8] by organizing these into a hierarchical structure. The gesture unit consists of a series of smallest observable primitives, phonemes: *preparation, stroke, retraction*. The *stroke* is distinguished by a peaking effort and it is thought to constitute the meaning of a gesture [8]. After extensive analysis of gestures in weather narration and *i*MAP [3, 4] we consider following strokes: *contour*, *point*, and *circle*. In addition to our previous definitions we also include *hold* as a functional primitive.

## 2.2. Synchronization of Gestures and Speech

The issue of how gestures and speech relate in time is critical for understanding the system that includes gesture and speech as part of a multimodal expression. At the phonological level, Kendon [8] found that different levels of movement hierarchy are functionally distinct in that they synchronize with different levels of prosodic structuring of the discourse in speech. E.g., stroke of the gesture was found to precede or end at the phonological peak syllable.

Both psycholinguistic, e.g., [6] and HCI, e.g., *i***MAP** [7], studies suggest that deictic gestures do not exhibit one-to-one mapping of form to meaning. Previously, we showed that the semantic categories of strokes (derived through the set of keywords), not the gesture phonemes, correlate with the temporal alignment of keywords, cf. [7]. This work distinguishes two types of gestures: referring to a static point on the map and to a moving object (i.e., moving precipitation front). Due to the homogeneity of the context and trained narrators in the weather domain we can statistically assume (mismatch <2%) that pointing gesture is the most likely to refer to the static and contour stroke to the moving objects. Therefore, for simplicity we will use *contour* and *point* definitions.

## 3. PROSODY BASED CO-ANALYSIS

The purpose of the current analysis is to establish a framework by identifying correlate features in visual and acoustic signals. First we will separate acoustically prominent segments. Segment is defined as a voiced interval on the pitch contour that phonologically can vary from a foot[1] to intonational phrase units, see [9] for details. Then we will analyze alignment of the prominent segment with the gesture phonemes.

Over 60 minutes of the weather narration data was used in the analysis. Every video sequence contained uninterrupted monologue of 1-2 minutes in length. The subject pool was presented by 5 men and 3 women.

### 3.1. Prosodic Features

By prominent segments are defined as segments which are relatively accentuated (or perceived as such) from the rest of the monologue. We consider combination of the pitch accent and the pause before each voiced segment to detect abnormalities in spoken discourse.

Pitch accent association in English underlines the discourse-related notion of focus of information. Fundamental frequency ($F_0$) is the correlate of pitch defined as the time between two successive glottis closures [10]. $F_0$ is considered to correlate with both a particular phones and overall tonal contrast, e.g., [11].

We employ an autocorrelation method that performs acoustic periodicity detection to extract $F_0$, as described in [12]. The resulted contour was pre-processed such that unvoiced intervals of less than 0.02 sec and less then 10Hz were interpolated between the neighboring segments.

### 3.2. Detecting Prominent Segments

We consider a problem of prominent segments detection to be a binary classification problem. We assume that prosodic prominence can be presented by a multidimensional feature set $\mathbf{f} = \left[\xi_{max}, \xi_{min}, \dot{\xi}_{max}\right]^T$, where $\mathbf{f} \sim \mathcal{N}(\mathbf{\mu}, \mathbf{\Sigma})$, $\xi_{max}$ and $\xi_{min}$ are accent measures, and $\dot{\xi}_{max}$ is the maximum gradient of the $F_0$ segment.

Accents $\xi_{max}$ is calculated as a product of the preceding pause duration and the $F_0$ differential between the end of the previous contour and the maximum of the current $F_0$ (Figure 1). Similarly we compute $\xi_{min}$ taking the minimum of $F_0$ contour. Max and min represent low and high pitch accents, c.f. [13].

To extract the maximum gradient of a pitch segment, $\dot{\xi}_{max}$, we used Canny's edge detection algorithm with a Gaussian smoothing ($\sigma$=0.8), for details see [14].

The solution for prominent $F_0$ segments detection are to be sought towards the "tails" of the Gaussian (Figure 2). Analysis of the constructed histograms indicated heavy tails for $\xi_{max}, \xi_{min}$ and $\dot{\xi}_{max}$ distributions. We applied Yeo-Johnson log transform [15] to improve normality.

---

[1] Foot is a phonological unit that has a "heavy" syllable followed by a "light" syllable(s).

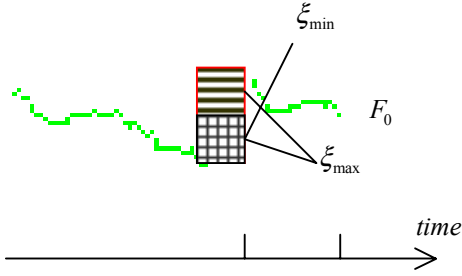

Figure 1. Calculation of accents for $F_0$ segment

*4.2.1. Prominence Detection Threshold*
To find an appropriate level of threshold to classify the segments we employed a bootstrapping technique involving a perceptual study. A control sample set for every narrator was labeled by 3 naïve coders for auditory prominence. The coders had access only to the wave form of speech signal.

Mahalanobis distance measure $d^2$ was used to define a decision boundary for prominent observations as labeled by the coders. Allowing 2% of misses, results indicated user dependent threshold ($d^2$=0.7-1.1). Figure 2 presents a sample distribution of **f** for a male narrator.

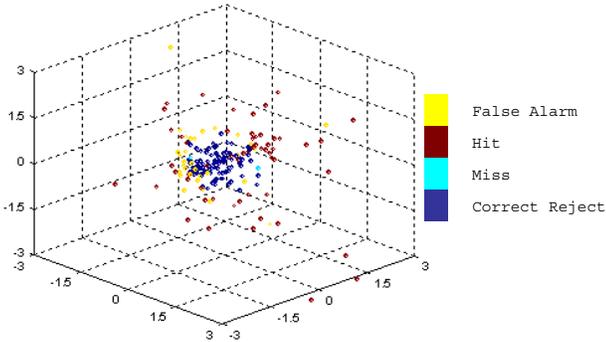

Figure 2. A sample distribution of **f** for a male with the decision boundary from the perceptual study.

We used results from the auditory perception study to bootstrap the current feature set **f** to form a meaningful clusters which consequently help in finding a robust solution.

### 3.3. Co-occurrence Model

We assume that temporal alignment of active hand velocity profile and prominent pitch segments for gesture classes $\omega_i$, where $i = 0,...n$, are distributed in a multidimensional feature space $\mathbf{t} = [\bar{\tau}_0, \bar{\tau}_{max}, \bar{\tau}_{min}, \bar{\tau}_{max'}]^T$, with Gaussian densities $\mathcal{N}_i(\mathbf{\mu}, \mathbf{\Sigma})$.

The gesture onset $\tau_0$, is the time from the beginning of the phoneme to the beginning of the prominent segment, which has at least a part of its $F_0$ contour presented within the duration of the gesture primitive (Figure 3).

The gesture peak onset $\tau_{max}$, is the time from the peak of the hand velocity to the maximum on the $F_0$ segment (high pitch accents). Similarly, we compute $\tau_{min}$ for low pitch accents (Figure 3).

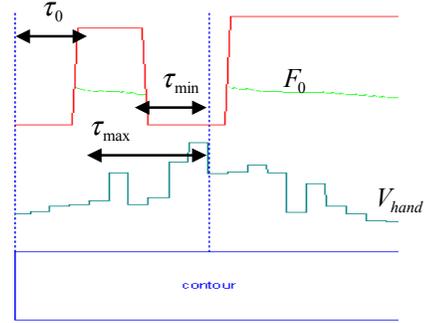

Figure 3. Co-occurrence features ($\tau_{max'}$ is not shown)

$\tau_{max'}$ is onset of maximum gradients of gesture velocity, $\dot{V}_{max}$, and $F_0$ segment, $\dot{\xi}_{max}$. $\dot{V}_{max}$ was found using Canny's formulation with $\sigma$=1.0. Note that **t** is a weighted average based on length of each $F_0$ segment, since a several $F_0$ segments might be present within the gesture phoneme.

All of 446 phonemes that has been used for training gesture phonemes were utilized for training of the co-occurrence models. Analysis of the resulted models indicated that there was no significant difference between retraction and preparation phases. Peaks of contour strokes tend closely to coincide with the peaks of pitch segments. Pointing appeared to be quite silent, the most of the segments were aligned with the beginning parts of the post-stroke hold.

## 5. CONTINUOUS GESTURE RECOGNITION

A gesture phoneme was defined as a stochastic process in the model parameter space **g** over suitably defined time interval. **g** is a seven element vector composed of velocities and accelerations from the blobs of the hand and the head. We employ HMM framework for continuous gesture recognition, as described in [3].

The fusion of the visual (gesture) and the co-occurrence models utilizes Bayesian framework. The likelihood $p(\mathbf{g}|\omega_i)$ is computed from **g** using Viterbi algorithm as described in **[3]**. The prior probability $P(\omega_i)$ is obtained from the joint signal of co-occurrence model. Then, the probability of observing phoneme of class $\omega_i$ becomes:

$$P(\omega_i|\mathbf{g}) \propto p(\mathbf{g}|\omega_i) P(\omega_i)$$

### 5.1. Results of Gesture Recognition

Table 1 presents summary of the recognition results. Total of 1876 gestures were segmented. The co-occurrence model yielded significantly better performance overall. High deletion rate in gesture-only case can be attributed to relatively short/small movement of the narrator that was assumed to belong one of the neighboring gesture phonemes. There was a significant improvement in reduction of deletion and substitution errors with co-occurrence model. It is mostly due to the inclusion of small point gestures which are quite salient when correlated with prominent acoustic features.

Table 1. Gesture Recognition Results in %

| Recognition mode | Hits | Deletion | Subst. | Insert |
|---|---|---|---|---|
| Gesture only | 72.4 | 16.1 | 9.2 | 2.3 |
| Gesture & Co-occurrence | 81.8 | 8.6 | 5.8 | 3.7 |

Figure 3 shows example of elimination of a substitution error due to the pointing stroke co-occurrence with a prominent spoken segment. Inclusion of the co-occurrence

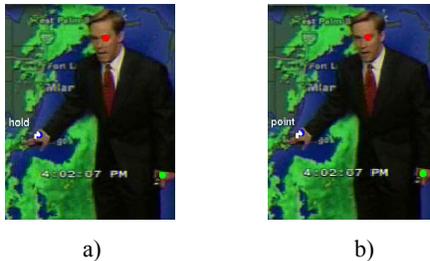

a)                  b)
Figure 3. Example of substitution error using: a)visual-only signal; b) with co-occurrence model. White trace presents hand movement trajectory.

models almost did not contribute to the improvement in insertion errors.

### 6. CONCLUSIONS

We presented an alternative approach for combining gesture and speech signals from the bottom-up perspective. Such formulation is more favorable for automated recognition of continuous co-verbal gestures then the semantic based (keyword co-occurrence). The current results confirms possibility of improving recognition of co-verbal gestures when combined with the prosodic features in speech. This is a first attempt which requires further improvement in formulation.